\documentclass[10pt,journal,compsoc]{IEEEtran}
\usepackage[noend]{algpseudocode}
\usepackage{bbm}
\usepackage{caption}
\usepackage{subcaption}
\usepackage{algorithm}
\usepackage{amssymb}
\usepackage{amsfonts}
\usepackage{amsmath}
\usepackage{amsthm}
\usepackage{graphicx}
\usepackage{mathbbol}
\graphicspath{{figures/}}
\usepackage[normalem]{ulem}
\newtheorem{theorem}{Theorem}

\newtheorem{corollary}{Corollary}
\newtheorem{prop}{Proposition}

\newtheorem{lemma}{Lemma}

\makeatletter
\def\BState{\State\hskip-\ALG@thistlm}
\makeatother
\usepackage{tikz}
\usetikzlibrary{arrows, automata}
\DeclareMathOperator*{\argmin}{arg\,min}
\DeclareMathOperator*{\argmax}{arg\,max}
\usepackage{multirow}
\usepackage{caption}
\usepackage{array}

\usepackage{xr}
\makeatletter
\newcommand*{\addFileDependency}[1]{
  \typeout{(#1)}
  \@addtofilelist{#1}
  \IfFileExists{#1}{}{\typeout{No file #1.}}
}
\makeatother

\newcommand*{\myexternaldocument}[1]{%
    \externaldocument{#1}%
    \addFileDependency{#1.tex}%
    \addFileDependency{#1.aux}%
}
\myexternaldocument{supplement}
\usepackage{cite}

\ifCLASSINFOpdf
\else
\fi
\begin{document}
%
\title{Learning Bayesian Networks through Birkhoff Polytope: A Relaxation Method}
%
%
%

\author{Aramayis~Dallakyan,
        Mohsen~Pourahmadi
\thanks{Department of Statistics, Texas A\&M University, College Station, TX, 77843}
}

\IEEEtitleabstractindextext{%
\begin{abstract}
We establish a novel framework for learning a directed acyclic graph (DAG) when data are generated from a Gaussian, linear structural equation model. It consists of two parts: (1) introduce a permutation matrix as a new parameter within a regularized Gaussian log-likelihood to represent variable ordering; and (2) given the ordering, estimate the DAG structure through sparse Cholesky factor of the inverse covariance matrix. For permutation matrix estimation, we propose a relaxation technique that avoids the NP-hard combinatorial problem of order estimation. Given an ordering, a sparse Cholesky factor is estimated using a cyclic coordinatewise descent algorithm which decouples row-wise. Our framework recovers DAGs without the need for an expensive verification of the acyclicity constraint or enumeration of possible parent sets. We establish numerical convergence of the algorithm, and consistency of the Cholesky factor estimator when the order of variables is known. Through several simulated and macro-economic datasets, we study the scope and performance of the proposed methodology.
\end{abstract}

}

\maketitle

%

\IEEEdisplaynontitleabstractindextext
\IEEEpeerreviewmaketitle

\section{Introduction}  \label{s:s1}
%
%
%
%
\IEEEPARstart{B}{ayesian} Networks (BNs) are a popular class of graphical models whose structure is represented by a DAG $\mathcal{G}$. BNs have been used in many applications such as economics,
 finance, biology, etc  \cite{Swanson1997, neil2005, needham2007, hyvarinen2010, dallakyan2020}. 
 In recent years the following two approaches have been evolved to learn the structure of the underlying DAG from data: Independence-based (also called constraint-based) methods \cite{spirtes1991, pearl2009} and score-based methods \cite{heckerman1995, Chickering2002, teyssier2005, loh2014}. Here, structure learning refers to recovering DAG from observational data.

Independence-based methods, such as the inductive causation (IC) \cite{pearl2009} and  PC (Peter-Clark) \cite{spirtes1991} algorithm, utilize conditional independence tests to detect the existence of edges between each pair of variables. The method assumes that the distribution is Markovian and faithful with respect to the underlying DAG, where $\mathcal{P}$ is faithful to the DAG $\mathcal{G}$ if all conditional independencies in $\mathcal{P}$ are entailed in $\mathcal{G}$ and Markovian if the factorization property (\ref{eq:markov}) is satisfied.

In contrast, score-based methods measure the goodness of fit of different graphs over data by optimizing a score function with respect to the unknown (weighted) adjacency matrix $B$ with a combinatorial constraint that the graph is DAG. Then a search procedure is used to find the best graph. Commonly used search procedures include hill-climbing \cite{heckerman1995, tsamardinos2006}, forward-backward search \cite{Chickering2002}, dynamic, and integer programming \cite{silander2006, koivisto2006, jaakkola10, studeny2014,hemmecke2012}. Recently, \cite{zheng2018, zheng2020} proposed a fully continuous optimization for structure learning by introducing a novel characterization of acyclicity constraint.

Generally, the DAG search space is intractable for a large number of nodes $p$ and the task of finding a DAG is NP-hard \cite{Chickering2002}. Consequently, approximate methods have been proposed with additional assumptions such as bounded maximum indegree of the node \cite{cooper1992} or tree-like structures  \cite{chow1968}. Alternatively, the ordering space (or the space of topological ordering) has been exploited for score-based methods \cite{teyssier2005, vandegeer2013,aragam2015, ye2019} where the topological ordering is considered as a parameter \cite{teyssier2005}. 
The order-based search has two main advantages: the ordering space ($2^{O(p\log p)}$) is significantly  smaller than the DAG search space ($2^{O(p^2)}$), and the existence of ordering guarantees satisfaction of the acyclicity constraint.

The recent Annealing on Regularized Cholesky Score (ARCS) algorithm in \cite{ye2019} is based on representing an ordering by the corresponding permutation matrix $P$, and  then given the order, encoding the weighted adjacency matrix $B$ into the Cholesky factor $L$ of the inverse covariance matrix.  ARCS optimizes a regularized likelihood score function to recover sparse DAG structure and utilizes simulated annealing (SA) to search over the permutation matrix space. In SA, using a pre-specified constant $m$ and a temperature schedule $\{T^{(i)}, i = 0, \dots, N\}$, in the $i$th iteration a new permutation matrix $P^{*}$ is proposed by flipping a fixed-length $m$ random interval in the current permutation $\hat P$, and checking whether to stay at the current $\hat P$ or move to the proposed $P^{*}$ with some probability.

Motivated by the ARCS two-step framework, we propose an order-based method for learning Gaussian DAGs by optimizing a non-convex regularized likelihood score function with the following distinct features and advantages:

First,  we use a relaxation technique instead of the expensive search for a permutation matrix $P$ in the  non-convex space of permutation matrices. More precisely, we project $P$ onto the Birkhoff polytope (the convex space of doubly stochastic matrices) and then find the ``closest'' permutation matrix to the optimal doubly stochastic matrix (See Figure~\ref{f:geom}). 
Second, given $P$, we resort to the cyclic coordinatewise algorithm to recover the DAG structure entailed in the Cholesky factor $L$. We show that the optimization reduces to $p$ decoupled penalized regressions where each iteration of the cyclic coordinatewise algorithm has a closed form solution. 
Third, we show consistency of our Cholesky factor estimator for the non-convex score function when the true permutation matrix is known. To the best of our knowledge, consistency results for the sparse Cholesky factor estimator were established only for convex problems \cite{yu2017, khare2016}.

The paper is organized as follows: Section 2 introduces background on Gaussian BNs and structural equation models (SEMs). In Section 3, we derive and discuss the form of the score function. In Section 4, we introduce our Relaxed Regularized Cholesky Factor (RRCF) framework. The analyses of the simulated and real macro-economic datasets are contained in Section 5. For the real data analysis, we apply RRCF to solve the price puzzle, a classic problem in the economics literature. Section 6 provides statistical consistency  of our estimator, and we conclude with a discussion in Section 7.

\section{Bayesian Networks} \label{s:bn}

We start by introducing the following graphical concepts. If the graph $\mathcal{G}$ contains a directed edge from the node $k \rightarrow j$, then $k$ is a parent of its child $j$. We write $\Pi^{\mathcal{G}}_{j}$ for the set of all parents of a node $j$. If there exist a directed path $k \rightarrow \dots \rightarrow j$, then $k$ is an ancestor of its descendant $j$. 
A \textit{Bayesian Network} is a directed acyclic graph $\mathcal{G}$ whose nodes represent random variables $X_1, \dots, X_p$. Then $\mathcal{G}$ encodes a set of conditional independencies   and conditional probability distributions for each variable. The DAG $\mathcal{G} = (V,E)$ is characterized by the node set $V = \{1, \dots, p\}$ and the edge set $E = \{(i,j): i \in \Pi^{\mathcal{G}}_j\} \subset V \times V$. It is well-known that for a BN, the joint distribution factorizes as:

\begin{equation} \label{eq:markov}
P(X_1,\dots,X_p) = \prod_{j=1}^{p}P(X_j| \Pi^{\mathcal{G}}_{j})
\end{equation}

 \subsection{Gaussian BN and Structural Equation Models}

    It is known that a Gaussian BN can be equivalently represented by the linear SEM \cite{pearl2009}:
 \begin{equation} \label{eq:sem}
 X_j = \sum_{k \in \Pi^{\mathcal{G}}_j} \beta_{jk} X_k + \varepsilon_j, \; j = 1, \dots, p,
 \end{equation}
 where $\varepsilon_j \sim N(0, \omega^2_j)$ are mutually independent and independent of $\{X_k: k \in \Pi^{\mathcal{G}}_j \}$. Denoting $B = (\beta_{jk})$ with zeros along the diagonal, the vector representation of (\ref{eq:sem}) is
 \begin{equation} \label{eq:vsem}
 X = BX + \varepsilon,
 \end{equation}
 where $\varepsilon: = (\varepsilon_1, \dots, \varepsilon_p)^t$ and $X: = (X_1, \dots, X_p)^t$.
Thus, one can characterize the linear SEM $X \sim (B, \Omega)$ by the weighted adjacency matrix $B$ and the noise variance matrix $\Omega = \mbox{diag}(\omega^2_1, \dots, \omega^2_p)$. From (\ref{eq:vsem}), the inverse covariance matrix of $X \sim N_p(0, \Sigma)$ is  $\Sigma^{-1} = (I-B)^t \Omega^{-1} (I-B)$,
 and the edge set of the underlying DAG is equal to the support of the weighted adjacency matrix $B$; i.e., $E = \{(k,j): \beta_{jk} \neq 0\}$, which defines the structure of DAG $\mathcal{G}$. Consequently, $B$ should satisfy the acyclicity constraint so that $\mathcal{G}$ is indeed a DAG.

\begin{figure}[t]
\centering
\includegraphics[width= 9cm, height = 5cm]{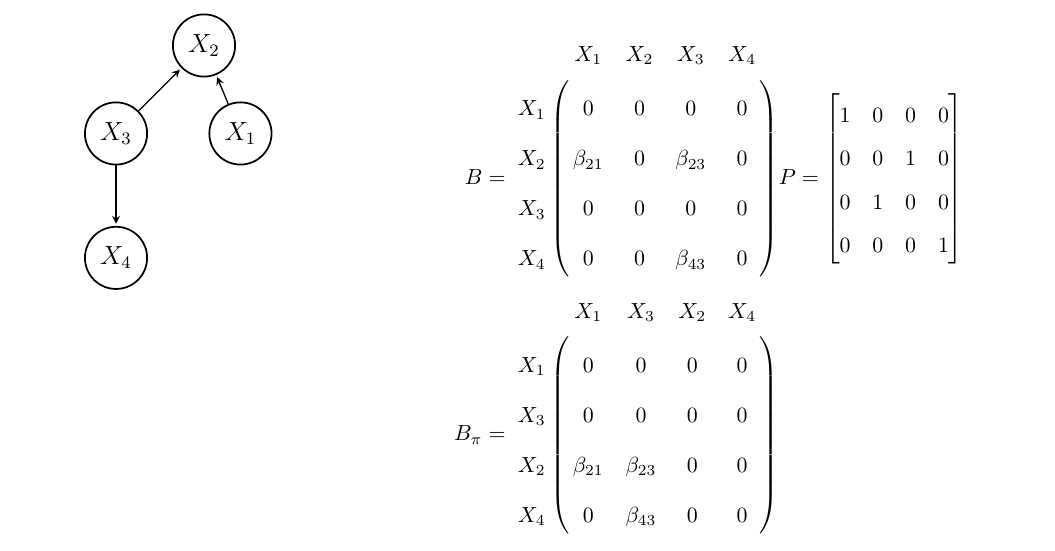}
\caption{Illustration of DAG $\mathcal{G}$, corresponding coefficient matrix $B$, permutation matrix $P$, and permuted strictly lower triangular matrix $B_{\pi}$.}
\label{fig:exdag}
\end{figure}

      It is known that a DAG admits a topological ordering $\pi$,  to which one may associate a $p \times p$ permutation matrix $P_{\pi}$ such that $P_{\pi}x = (x_{\pi(1)}, \dots, x_{\pi(p})$, for $x \in R^p$. The existence of a topological order leads to the permutation-similarity of $B$ to a strictly lower triangular matrix $B_{\pi} = P_{\pi}BP^t_{\pi}$ by permuting rows and columns of $B$, respectively \cite{bollen1989} (see Figure~\ref{fig:exdag} for the illustrative example). Therefore, the stringent acyclicity constraint on $B$ transforms into the constraint that $B_{\pi}$ is a strictly lower triangular matrix, then the linear SEM can be rewritten as
 \begin{equation} \label{eq:pvsem}
  P_{\pi}X = B_{\pi} P_{\pi}X + P_{\pi}\varepsilon,
  \end{equation}
 using the fact that $P^t_{\pi}P_{\pi} = I$. From (\ref{eq:pvsem}),  the inverse covariance matrix can be expressed as
  \begin{equation} \label{eq:pcov}
  \Sigma^{-1}_{\pi} = (I - B_{\pi})^t \Omega^{-1}_{\pi} (I - B_{\pi}),
  \end{equation}
  where $\Omega_{\pi} = P_{\pi}\Omega P^t_{\pi}$.
  Using~(\ref{eq:pvsem}) and (\ref{eq:pcov}) and defining $L_{\pi} = \Omega_{\pi}^{-1/2}(I - B_{\pi})$,  the relationship between the Cholesky factor $L_{\pi}$ of the inverse covariance matrix $\Sigma^{-1}_{\pi}= L^t_{\pi}L_{\pi}$ and the matrix $B_{\pi}$ is
   \begin{equation} \label{eq:lb}
   \begin{split}
   (L_{\pi})_{ij} &= -(B_{\pi})_{ij} / \sqrt{\omega_{j}},\;\mbox{and}\;\\
     (L_{\pi})_{ij} &= 0    \iff (B_{\pi})_{ij} = 0\; \mbox{for every}\; i \geq j
   \end{split}
   \end{equation}
Hence, $L_{\pi}$ preserves the DAG structure of $B_{\pi}$; i.e., non-zero elements in $L_{\pi}$ correspond to directed edges in DAG $\mathcal{G}$. 

\section{The Score Function} \label{s:rrcf}

In this section, given data from the Gaussian BN (or SEM), we derive the form of the score function used to recover the underlying DAG structure. A natural choice for such function is the log-likelihood function, which will be used for the estimation of the permutation and Cholesky factor matrices.
We assume that each row of  data matrix $\mathbf{X} = (X_1, \dots, X_p) \in R^{n \times p}$  is an i.i.d observation from (\ref{eq:sem}). Using reformulation (\ref{eq:pvsem}),
\begin{equation} \label{eq:pvsemmat}
\mathbf{X}P_{\pi}^t = \mathbf{X} P_{\pi}^t B_{\pi}^t + \mathbf{E}P_{\pi}^t,
\end{equation}
where each row of $\mathbf{E}$ is an i.i.d $N_p(0, \Omega)$ vector. Thus, each row of $\mathbf{X}P_{\pi}^t$ is, again, an i.i.d from  $N_p(0, \Sigma_{\pi})$, and the negative log-likelihood for (\ref{eq:pvsemmat}) is:
\begin{equation} \label{eq:loglik}
\begin{split}
\ell (B_{\pi}, \Omega_{\pi}, P_{\pi} | \mathbf{X}) &= \frac{1}{2} \mbox{tr} \Big (P_{\pi} \mathbf{X}^t\mathbf{X}P_{\pi}^t(I - B_{\pi})^t \Omega_{\pi}^{-1}(I - B_{\pi}) \Big ) \\
& + \frac{n}{2}  \log |\Omega_{\pi}|,
\end{split}
\end{equation}
using the facts that $\Sigma_{\pi} = \mbox{cov}(P_{\pi}X) = P_{\pi}\Sigma P^t_{\pi} = (I - B_{\pi})^{-1} \Omega_{\pi} (I - B_{\pi})^{-t}$, and $B_{\pi}$ is a strictly lower triangular matrix. 
 From now on, whenever there is no confusion, we drop the subscript $\pi$ from $P_{\pi}, B_{\pi}, \Omega_{\pi}$ and $\Sigma_{\pi}$.
%

After reparametrizing (\ref{eq:loglik}) in terms of $L$ and $P$, it reduces to
\begin{equation} \label{eq:parloglik}
\ell (L, P | \mathbf{X}) = \frac{1}{2} \mbox{tr} \Big (P SP^tL^t L \Big ) - \sum_{j = 1}^p\log L_{jj},
\end{equation}
where  $S = \mathbf{X}^t \mathbf{X} / n$ is the sample covariance and  $|\Omega|^{-1/2}= |L| = \prod_{j =1}^p L_{jj}$. Unfortunately, as stated in the next proposition, $\ell(L,P| \mathbf{X})$ is permutation invariant and maximum likelihood does not favor any particular ordering. Consequently, all maximum likelihood DAGs corresponding to a different permutation produce the same value of the Gaussian log-likelihood function.

\begin{prop} \label{p:prop1}
	The log-likelihood function  $\ell(L,P| \mathbf{X})$, defined in (\ref{eq:loglik}),  is permutation invariant, i.e., $\ell(L, I| \mathbf{X}) = \ell(L, P| \mathbf{X})$, where $I$ is the $p \times p$ identity matrix.
\end{prop}

We note that \cite[Proposition 1]{ye2019} provide a similar result, but a closer look at their proof  reveals that it is valid only for the (lower dimensional) case $n > p$. Our result is more general and its proof can be utilized to verify the permutation invariance for other score functions. The key idea in our proof is showing Schur-convexity \cite[Chapter 3]{marshall11} of the score function.


We follow \cite{ye2019} to break the permutation invariancy in (\ref{eq:loglik}) and regularize the negative log-likelihood function to favor sparse DAGs, hence learning a better model \cite[Chapter 18.1]{koller2009}.  We consider the following penalized score function:

\begin{equation} \label{eq:frrcf}
\begin{split}
 \min_{L \in \mathcal{L}_p ,P \in \mathcal{P}_p}Q(L,P) &=  \min_{L,P} \Big \{\frac{1}{2} \mbox{tr} \Big (P SP^tL^t L \Big ) \\ &- \sum_{j = 1}^p\log L_{jj}\\ & + \sum_{1 \leq j \leq i \leq p} \rho(|L_{ij}|; \lambda) \Big \},
 \end{split}
\end{equation}
where $\mathcal{P}_p,\, \mathcal{L}_p$  are the sets of all $p \times p$ permutation and lower triangular matrices with positive diagonal entries and the penalty function $\rho(\cdot,\lambda): \mathcal{R} \rightarrow \mathcal{R}$ satisfies conditions listed in \cite{loh2015} and reiterated in Supplementary for convenience. These conditions are required for establishing theoretical properties of our estimators in Section~\ref{s:statprop}.



\section{A minimization algorithm}

In this section, we introduce our two-step algorithm to minimize the score function (\ref{eq:frrcf}), named Relaxed Regularized Cholesky Factor (RRCF).  First, we propose a  relaxation  to solve the optimization problem in line 5 through a gradient projection algorithm (see Algorithm~\ref{a:ppm}). Then estimate a Cholesky factor in  line 6 utilizing a cyclic coordinatewise algorithm (see Algorithm \ref{a:cca}). We show that in the first step, a convex relaxation can be achieved when the number of observations exceeds the number of variables. 

\begin{algorithm}[ht!]
\caption{RRCF algorithm}\label{a:RRCF}
\begin{algorithmic}[1]
\BState \emph{input}:
\State $\textit{$\lambda,k_{max}$} \gets \textit{Tuning Parameter, iteration}$
\State $\textit{${L}^{(0)}, P^{(0)}$} \gets \textit{Initial matrices }$
\BState \emph{while $k < k_{max}$}: 
\State   $\quad \quad  \hat P^{(k)} = \argmin_{P \in \mathcal{P}_p} Q_{RRCF}(L^{(k - 1)}, P) $  \label{a:SSC:1} 
\State   $\quad \quad  \hat L^{(k)} = \argmin_{L \in \mathcal{L}_p} Q_{RRCF}(L, P^{(k)}) $  \label{s:hi}
\State  $ \quad \quad k = k+1$
\BState \emph{Output}:$\;(\hat L, \hat P)$
\end{algorithmic}
\end{algorithm}

\subsection{Optimization over the permutation space}
A paramount issue in finding an optimal permutation matrix is that the size of the search space is $p!$.
\cite{ye2019} mitigate the problem by using a simulated annealing technique to search over the permutation space. Our approach is significantly different and relies on enlarging the non-convex set of permutation matrices to the convex set of doubly stochastic matrices (Birkhoff polytope) and finding the ``closest'' permutation matrix to the optimal doubly stochastic matrix. In view of the recent advances in Seriation \cite{fogel2013} and Graph Matching problems \cite{zaslavskiy2009,wolstenholme2016}, our approach amounts to  a  relaxation of the hard combinatorial problem.

\subsubsection{A Convex Relaxation}
The impetus of the work in this section is  the framework in \cite[Section 3.2]{fogel2013}. 
 The optimization in line 5 of Algorithm~\ref{a:RRCF} can be written as:
\begin{equation} \label{p:perm}
\begin{aligned}
\min_{P} \quad & \frac{1}{2}\mbox{tr}(LPSP^tL^t) \\
\mbox{s.t.} \quad & P \in \mathcal{P}_p ,
 \end{aligned}
\end{equation}
where we eliminate terms that are constant with respect to $P$. We denote the Birkhoff polytope by $\mathcal{D}_p$  (the space of doubly stochastic matrices), where $\mathcal{D}_p = \{A \in R^{p \times p}: A \geq 0, A \mathbf{1} = \mathbf{1}, A^t \mathbf{1} = \mathbf{1}\}$, it has $p!$ vertices and dimension of $(p - 1)^2$. It is informative to note that every permutation matrix is a doubly stochastic matrix, and a matrix is a permutation if and only if it is both doubly stochastic and orthogonal; i.e., $\mathcal{P}_p = \mathcal{D}_p \cap \mathcal{O}_p$, where $\mathcal{O}_p$ is the set of $p \times p$ orthogonal matrices. Moreover, from Birkhoff's Theorem, every doubly  stochastic matrix can be written as a convex combination of permutation matrices and the set of doubly stochastic matrices is the convex hull of the set of permutation matrices \cite[Theorem 8.7.2]{Horn2012}, where permutation matrices are vertices (extreme points) of the polytope. More on Birkhoff polytopes and its properties can be found in \cite{brualdy1974}.

Since the sample covariance matrix $S \succcurlyeq 0 $ is positive semi-definite, we can introduce a convex relaxation to the combinatorial problem~(\ref{p:perm}) by replacing $\mathcal{P}_p$ with its convex hull $\mathcal{D}_p$:

\begin{equation} \label{p:stoc}
\begin{aligned}
\min_{P} \quad & \frac{1}{2}\mbox{tr}(LPSP^tL^t) \\
\mbox{s.t.} \quad & P \in \mathcal{D}_p ,
 \end{aligned}
\end{equation}

However, as  shown in Corollary~\ref{c:optsol}, the solution of (\ref{p:stoc}) is not an acceptable candidate. In the next lemma, we list well-known properties of doubly stochastic matrices that are used to establish the framework for the convex relaxation. Since  we are not aware of a source to cite, a proof is given in the Supplementary for completeness. We denote by $J \in \mathcal{D}_p$ the $p \times p$ matrix all of whose entries are $1$.
\begin{lemma} \label{l:dsprop}
For any $p \times p$ doubly stochastic matrix $P \in \mathcal{D}_p$,
\item \[1 \leq \|P\|_F \leq \sqrt{p}\]
The left and right equalities hold if and only if $P = J/p$ and $P$  is a permutation matrix, respectively.
\end{lemma}

From Lemma~\ref{l:dsprop}, the following corollary easily follows.

\begin{corollary} \label{c:optsol}
The optimal solution of (\ref{p:stoc}) is $ \hat P = J/p$.
\end{corollary}

\begin{figure}[t]
\centering
\resizebox{4cm}{3cm}{%
\begin{tikzpicture}
   \newdimen\R
   \R=2.7cm
   \node[inner sep = 0.5pt, circle, draw] {J/p};
   \draw (0:\R) \foreach \x in {60,120,...,360} {  -- (\x:\R) };
   \foreach \x/\l/\p in
     { 60/{(2,3,1)}/above,
      120/{(2,1,3)}/above,
      180/{(1,2,3)}/left,
      240/{(1,3,2)}/below,
      300/{(3,1,2)}/below,
      360/{(3,2,1)}/right
     }
     \node[inner sep=1pt,circle,draw,fill,label={\p:\l}] at (\x:\R) {};
\end{tikzpicture}%
}
\caption{A geometric depiction of relaxation (\ref{p:stocort}) for $\mathcal{D}_3$ Birkhoff polytope. Here, vertices represent permutations, and matrix $J/p$ indicates the center of the polytope.}
\label{f:geom}
\end{figure}
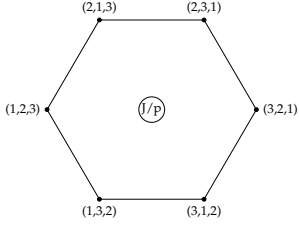

Thus, the solution of  (\ref{p:stoc}) is the center of the Birkhoff polytope  \cite[page 20]{ziegler1995} and far from vertices where permutation matrices are located.  To force it to move closer to the vertices,  we utilize Lemma~\ref{l:dsprop} to motivate and add a proper penalty to the objective function (See Figure~\ref{f:geom} for the geometric depiction.)
\begin{equation} \label{p:stocort}
\begin{aligned}
\min_{P} \quad & \frac{1}{2}\mbox{tr}(LPSP^tL^t)  - \frac{1}{2}\mu\|P\|^2_F\\
\mbox{s.t.} \quad & P \geq 0, P \mathbf{1} = \mathbf{1}, P^t \mathbf{1} = \mathbf{1}.
 \end{aligned}
\end{equation}
Note that for  larger $\mu > 0$, $\|P\|^2_F$ is pushed toward its upper bound  $p$ (Lemma~\ref{l:dsprop}), so that the larger $\mu$, the closer the solution of (\ref{p:stocort}) is to a permutation matrix. 
Similar to \cite[Proposition 3.5]{fogel2013}, the next lemma shows that the convexity of the objective function (\ref{p:stocort}) depends on the inextricably intertwined values of $\mu$ and the smallest eigenvalue of $S$ and $L^tL$. As a result, the convexity is untenable when $n<<p$. We mitigate this problem by introducing an additional transformation to maintain convexity when $n \approx p$ (Lemma~\ref{l:l1}(b)). The following notation is used in the lemma: we write $\lambda_1 < \lambda_2 < \dots < \lambda_m$ as an ordered, distinct eigenvalues of the $p \times p$ matrix. The proof is provided in the Supplementary for completeness.

\begin{lemma} \label{l:l1}
\begin{enumerate}
\item[a.] If $\mu \leq  \lambda_1(S)\lambda_1(L^tL)$, the optimization problem  (\ref{p:stocort}) is convex in $P$.
\item[b.] If $\mu \leq \lambda_2(S)\lambda_1(L^tL)$  and $T = \mathbf{I} - \frac{1}{p} \mathbf{1}\mathbf{1}^t$ is the projection matrix into the orthogonal complement of $\mathbf{1}$, then the optimization problem
\begin{equation} \label{p:trstoc}
\begin{aligned}
\min_{P} \quad & \frac{1}{2}\mbox{tr}(LPSP^tL^t)  - \frac{1}{2}\mu\|TP\|^2_F\\
\mbox{s.t.} \quad & P \geq 0, P \mathbf{1} = \mathbf{1}, P^t \mathbf{1} = \mathbf{1},
 \end{aligned}
\end{equation}
is equivalent to problem (\ref{p:stocort}) and is convex in $P$.
\item[c.] If $\mu > \lambda_m(S)\lambda_m(L^tL)$, the optimization problem (\ref{p:stocort}) is concave in $P$ and the solution is a permutation matrix.
\end{enumerate}
\end{lemma}


From Lemma~\ref{l:l1}(a) and (b), for $n <<p$,  $\lambda_2(S)$ is zero, and there is no $\mu >0$ that validates convexity of (\ref{p:trstoc}).
The question we investigate next is whether, under the convexity assumption of Lemma~\ref{l:l1}(a) or (b), there is a value of $\mu$ that asymptotically achieves ``closeness'' to the permutation matrix in terms of Frobenius norm.
We provide the answer only for  (\ref{p:stocort}), but the similar result holds for (\ref{p:trstoc}) by analogy. 
Recall that we tacitly assume the condition $n > p$ to maintain convexity.

\begin{lemma} \label{l:lbound}
Under convexity condition in  Lemma~\ref{l:l1}(a), for $\mu > 0$
\begin{equation} \label{e:lbound}
	\|\hat P - P\|_F \not \rightarrow 0,\, \mbox{as}\; n \rightarrow \infty,
\end{equation}
where $P \in \mathcal{P}_p$ and $\hat P$ is the solution of  (\ref{p:stocort}).
\end{lemma}

The proof can be found in the Supplementary. Lemma~\ref{l:lbound} suggests that under a convexity condition, the solution of   (\ref{p:stocort}) does not get ``close'' to the permutation matrix, even when $n \rightarrow \infty$.
The result may encourage the use of higher values of $\mu$, resulting in a non-convex objective function. However, this approach is not recommended. Our empirical results suggest that for comparably large $\mu$, the RRCF algorithm becomes independent from the data  and highly dependent on the initial choice of $P$.
Consequently, it gets stuck at one of the extreme points of the Birkhoff polytope.
The choice of $\mu$ for this setting is an open question and left for further investigation. Here, when $n <p$, we propose to treat $\mu$ as a tuning parameter and use information criteria or cross-validation for the selection.

%

\subsubsection{Gradient Projection Algorithm}

We provide details for solving (\ref{p:trstoc}), but the procedure similarly applies to (\ref{p:stocort}). Optimization (\ref{p:trstoc}) is a quadratic program (QP), and rich literature exists on solving this class of problems. 
 In this section, we rely on the Gradient Projection \cite{bertsekas2015} method and show the convergence of the algorithm. Algorithm~\ref{a:agp} outlines general steps, where $[\cdot]^+$ denotes projection on the space of doubly stochastic matrices $\mathcal{D}_p$.  


 \begin{algorithm}[ht!]
\caption{Gradient Projection}\label{a:agp}
\begin{algorithmic}[1]
\BState \emph{input}:
\State $\textit{$k_{max},\mu, \eta$} \gets \textit{the number of iterations and positive scalars}$
\State $\textit{$L, P^{(0)}$} \gets \textit{Cholesky and Initial Permutation matrix }$
\BState \emph{while $\|P^{(k+1)}- P^{(k)}\|>\epsilon$ }: 
\State   $\quad  \hat P^{(k+1)} = [P^{(k)} - \eta \nabla Q_{RRCF}(P^{(k)}, L) ]^+\;\;$ \textit{via Algorithm~\ref{a:dsm}}
\State   $\quad   P^{(k+1)} =  P^{(k)} + \alpha^k (\hat P^{(k+1)} -  P^{(k)})$
\State  $ \quad k = k+1$
\BState \emph{Output}:$\; \textit{Doubly Stochastic Matrix}\; P $
\end{algorithmic}
\end{algorithm}
 Line 5 of the algorithm requires projection onto the Birkhoff polytope, which can be efficiently implemented by the block coordinate ascent, where each iteration has a closed form solution. The details on the block coordinate ascent algorithm are given in the next section. The convergence of the algorithm to a global minimum easily follows from the \cite[Proposition 6.1.2]{bertsekas2015}.

%
\subsubsection{Projection onto the Birkhoff Polytope}

Here, we give details on solving line 5 of Algorithm~\ref{a:agp}. For a given matrix $P_0$, its projection onto $\mathcal{D}_p$ is defined by

\begin{equation} \label{p:proj}
\begin{aligned}
\min_{P} \quad & \frac{1}{2}\|P - P_o\|^2_F\\
\mbox{s.t.} \quad & P \geq 0, P \mathbf{1} = \mathbf{1}, P^t \mathbf{1} = \mathbf{1}.
 \end{aligned}
\end{equation}
The Lagrangian of (\ref{p:proj}) is  \cite{bertsekas2015}
\[
\begin{split}
\mathcal{L}(P,u,v,U) &= \frac{1}{2}\|P - P_0\|^2_F + u^t (P \mathbf{1} - \mathbf{1} )\\
&+ v^t(P^t \mathbf{1} - \mathbf{1})  - tr(U^t P) ,
\end{split}\]
and the dual objective function is defined as:
\begin{equation} \label{e:dual}
\mathcal{L}_{*}(u,v,U) = \inf_{P}\mathcal{L}(P,u,v,U)
\end{equation}
Consequently, the dual problem of (\ref{p:proj}) is (see Supplementary 
for details)
 \begin{equation} \label{p:dual}
\begin{aligned}
\max_{u,v,U} \quad & -\frac{1}{2}\|u\mathbf{1}^t + \mathbf{1}v^t - U\|^2_F - tr(U^tP_0)\\
& +u^t(P_0\mathbf{1} - \mathbf{1}) + v^t(P_0^t\mathbf{1} - \mathbf{1}) \\
\mbox{s.t.} \quad & U \geq 0 ,
 \end{aligned}
\end{equation}
Following \cite[Section 4.2]{fogel2013}, we use the block coordinate ascent algorithm to optimize the dual problem (\ref{p:dual}). We show that each block update has a closed form solution.
 Details of the algorithm and the derivation of closed form solutions are relegated to the Supplementary.

  In the next section, we propose a framework to find the  ``closest'' permutation matrix to the doubly stochastic matrix solution  (\ref{p:stocort}) or (\ref{p:trstoc}).

 \subsubsection{Sampling Permutations from the Space of Doubly Stochastic Matrices}

Since the solution of a convex relaxation (\ref{p:stocort}) is not a permutation matrix, we need to project it to the ``closest'' matrix $P \in \mathcal{P}_p$. 

Let $\tilde P$ be the doubly stochastic matrix solution of (\ref{p:stocort}), then a common method to project this matrix onto the  space of permutation matrices is through the following optimization \cite[Section 2.1]{zaslavskiy2009}:
\begin{equation} \label{e:hung}
\argmin_{P \in \mathcal{P}_p} \|\tilde P - P\|^2_F = \argmax _{P \in \mathcal{P}_p} tr \{\tilde P^t P\},
\end{equation}
which is a linear assignment problem and usually solved by the Hungarian algorithm \cite[Section 4.2.1]{burkard2012} and  takes $O(p^3)$ operations.

Unfortunately, (\ref{e:hung}) suffers a serious drawback as it only delivers one candidate solution to (\ref{p:stocort}), and if it is not ``close'' to the true permutation matrix $P$, it is unclear how to continue \cite[Section 3]{wolstenholme2016}. A viable alternative is a  permutation sampling procedure initially proposed for the orthogonal matrices in \cite{barvinok2005}. The idea is to ``round'' an orthogonal matrix $Q$ to a permutation matrix $P$ by considering its action on a random vector sampled from a Gaussian distribution. Consider a sample $x \in R^p$ from a Gaussian distribution and an ordering vector $r(x)$ such that $r(x)_i = k$ where $x_i$ is the $k$th smallest value of $x$. For example, if $x = [4.7,-2.1, 2.5 ]^t \Rightarrow r(x) =[3,1,2]^t $. Barvonik argues, if the permutation matrix $P$ satisfies
\begin{equation} \label{e:samp}
P(r(x)) = r(Qx)
\end{equation}
then it is ``close'' in Frobenious  norm to $Q$ with respect to $x$, as they both act on $x$ in a similar way  \cite[Theorem 1.6]{barvinok2005}.  In other words, $P$ matches the $k$th smallest coordinate of $x$ with the $k$th smallest coordinate of $Qx$, and $P$ represents a ``rounding'' of $Q$. This provides a framework to project an orthogonal matrix to a distribution of permutation matrices.

A close examination of  proof of \cite[Theorem 1.6]{barvinok2005} reveals that it is not restricted to orthogonal matrices and can be successfully extended to doubly stochastic matrices \cite[Section 4A]{wolstenholme2016}.
We use (\ref{e:samp}), selecting a doubly stochastic matrix $\tilde P$ instead of $Q$, to generate $N$ permutation matrices each ``close'' to the doubly stochastic matrix $\tilde P$. Then a common way to select the ``best'' permutation matrix from the $N$ sampled matrices is to pick a matrix that provides the lowest cost to (\ref{p:stoc}) \cite[Section 3.2.4]{fogel2013}.

Finally, Algorithm~\ref{a:ppm} combines necessary steps to estimate a permutation matrix $P$ in line 5: estimation of the doubly stochastic matrix (\ref{p:trstoc}),  and its approximation to the ``closest'' permutation matrix via (\ref{e:samp}).

 \begin{algorithm}[ht!]
\caption{Optimization over permutation matrices}
\label{a:ppm}
\begin{algorithmic}[1]
\BState \emph{input}:
\State $\textit{$N_{max}$} \gets \textit{max. number of sampling}$
\State $\textit{Find $\tilde P$ via Algorithm~\ref{a:agp}} $
\BState \emph{if $\tilde P \not \in \mathcal{P}_p$}:
\BState $\quad$ \emph{while $ j < N_{max}$}: 
\State $\quad \quad \textit{Sample:}\; x^{(j)} \sim N(0, \mathbf{I}_p)\}$
\State $\quad \quad  \textit{Solve for $P^{(j)}$ using (\ref{e:samp}):}$
\BState \emph{\textit{From $\{P^{(j)}\}_{j = 1}^{N_{max}}$ choose} $P$ \text{that minimizes (\ref{p:stoc}) }}.
\BState \emph{Output}:$\; \textit{Permutation Matrix}\; P $
\end{algorithmic}
\end{algorithm}
\subsection{Cholesky Factor Estimation}
This section focuses on Cholesky factor $L$ estimation from line 6 of Algorithm~\ref{a:RRCF}. That is we fix a permutation matrix $P$ and update the Cholesky factor $L$ using a \textbf{non-convex} objective function (\ref{eq:frrcf}). It is informative to recall that a Cholesky factor $L$ entails the DAG structure, and by learning $L$, accordingly, we learn the DAG structure in $B$.

Given an ordering, \cite{shojaie2010} and convex sparse Cholesky selection (CSCS) algorithm proposed in \cite{khare2016} estimate sparse Cholesky factor $L$ using a lasso-based penalty and convex objective function. 
Here, for the fix permutation matrix  $P$, we propose a cyclic coordinatewise algorithm to learn the Cholesky factor $L$ from the  non-convex objective function (\ref{eq:frrcf}). 
We show that the objective function can be decoupled into $p$ parallel penalized regression problems. The latter can be compared with the decomposable property of the score function in the BN literature, since non-zero values in each $i$th row of $L$ correspond to the parents of the $i$th node in DAG. Recall that a score function $f(\mathcal{G}, X)$ is decomposable if it can be written as $f(\mathcal{G}, X) = \sum_{i=1}^p f(X_i| \Pi^{\mathcal{G}}_{i})$ \cite[Definition 18.2]{koller2009}.

Denoting  $S^P = PSP^t$,\; $S^P_i$ the $i \times i$ sub-matrix of $S^p$, $L_{i \cdot}$ the $i$th row of $L$, and $\beta^i$ non-zero values of the $L_{i \cdot}$, it follows from (\ref{eq:frrcf})  

\begin{equation}
\begin{aligned}
Q_{RRCF}(L)& =
tr(LS^pL^t) - 2 \sum_{i=1}^{p}\log L_{ii}\\
  &+ \sum_{1\leq j < i \leq p} \rho(|L_{ij}|, \lambda) = \sum_{i = 1}^{p} (\beta^i)^t S^P_i \beta^i\\
  &- 2 \sum_{i =1}^p \log (\beta^i_i) 
  + \sum_{i =2}^{p} \sum_{j =1}^{i-1} \rho(|\beta^i_j|, \lambda)\\
 &= \sum_{i =1}^p Q_{RRCF,i}(\beta^i),
\end{aligned}
\end{equation}
where in arguments of $Q_{RRCF}(\cdot)$ we omit the dependence from $P$,  and
\begin{equation} \label{eq:rrcfi}
\begin{split}
Q_{RRCF,i}(\beta^i) &= (\beta^i)^t S^P_i \beta^i - 2 \log \beta^i_i \\ & + \sum_{j = 1}^{i -1} \rho(|\beta^i_j|, \lambda)
\end{split}
\end{equation}
for $2 \leq i \leq p$, and
\begin{equation}\label{eq:rrcf1}
 Q_{RRCF,1}(L_{11}) = L^2_{11}S^P_{11} - 2 \log L_{11}
 \end{equation}

In (\ref{eq:rrcfi}), we focus on the class of penalties called the minimax concave penalty (MCP) \cite{zhang2010}. MCP exploits convexity of the penalized loss near the sparse regions and concavity outside. It includes $\ell_1$ and $\ell_0$ as extreme cases and with two parameters $(\gamma,\lambda)$ takes the form

 \begin{equation}\label{eq:mcp}
  \rho(\theta, \lambda, \gamma) = \begin{cases} \lambda |\theta| - \frac{\theta^2}{2\gamma} & |\theta| < \gamma \lambda\\
  \frac{1}{2} \gamma \lambda^2 & |\theta| \geq \gamma \lambda,
  \end{cases}
  \end{equation}
  where $\lambda \geq 0$ and $\gamma >1$. 


Next, we derive steps to minimize the score function $Q_{RRCF}(L)$ with respect to non-zero values of $L$ for the fixed $P$. We assume that diagonal entries of the sample covariance matrix $S$ are strictly positive. Since $\{\beta^i\}^p_{i=1}$ disjointly partition the parameters in $L$, then optimizing $Q_{RRCF}(L)$ can be implemented by separately optimizing $Q_{RRCF}(\beta^i)$ for $1 \leq i \leq p$.

We define a generic function $h: R^{k-1} \times R_+ \rightarrow R$ of the form
\begin{equation} \label{e:gen}
h_{k,A,\lambda,\gamma}(x) = - 2 \log x_k + x^tAx  +  \sum_{i =1}^{k -1} \rho(|x_i|, \lambda, \gamma),
\end{equation}
  where $\lambda>0,\, \gamma > 1$ and $A$ is a positive semi-definite matrix with positive diagonal entries. It is instructive note that $Q_{RRCF,i}(\beta^i) = h_{i,S_i,\lambda,\gamma}(\beta^i)$ for every $1 \leq i \leq p$, and it suffices to develop an algorithm which minimizes a function of the form $h_{i,A,\lambda,\gamma}$. For every $1 \leq j \leq k$, we define
  \[x^*_j = \inf_{x_j}h_{k,A,\lambda,\gamma}(x).\]
   Next lemma shows that $\{x^*_j\}_{j=1}^{k}$ can be computed in the closed form. The proof is given in Supplementary.

   \begin{lemma} \label{l:xstar}
    The optimal solution $\{x^*_j\}_{j =1}^{k}$ can be computed in the closed form.
   \begin{equation} \label{eq:xk}
   x^*_k = \frac{-\sum_{l \neq k}A_{lk}x_l + \sqrt{(\sum_{l \neq k}A_{lk}x_l)^2 +4A_{kk}}}{2A_{kk}}
   \end{equation}
   and for $1 \leq j \leq k-1$,
   \begin{equation}\label{eq:xj}
   x^*_j = \frac{S_{\lambda}(-2\sum_{l \neq k}A_{lk}x_l)}{2A_{jj} - 1/\gamma}
   \end{equation}
   \end{lemma}
Here, $S_{\lambda}$ is the soft-thresholding operator given by $S_{\lambda}(x) = sign(x)(|x| - \lambda)_{+}$. From Lemma~\ref{l:xstar}, Algorithm~\ref{a:cca} provides a cyclic coordinatewise minimization algorithm for $h_{k,A,\lambda,\gamma}$.
We use it  to minimize $Q_{RRCF}(\beta^i)$ for $1 \leq i \leq p$, and combine outputs to obtain the estimated Cholesky factor $L$ in Algorithm~\ref{a:ecf}.

\begin{algorithm}[ht]
\caption{Cyclic coordinatewise algorithm}\label{a:cca}
\begin{algorithmic}[1]
\BState \emph{input}:
\State $\textit{$k_{max},A, \lambda, \gamma, \epsilon$}$
\State $x^{(0)} \gets \textit{Initial estimate} $
\State $\textit{Set}\; x^{\textit{current}} = x^{(0)};\;\mbox{Converged}=\mbox{FALSE} $
\BState \emph{while $\mbox{Converged} ==\mbox{FALSE}$ or $ k < k_{max}$ }: 
\State $\quad  x^{old} \gets x^{current}$
\State $\quad  \textit{For}\; $j = 1,2,\dots, k - 1$  $
\BState \emph{$\quad \quad x^{current}_j = x^*_j\;\;\textit{via (\ref{eq:xj})}$}
\State $\quad x^{current}_k = x^*_k\;\;\textit{via  (\ref{eq:xk})}$
\State $\quad \emph{if}\; \|x^{current} - x^{old}\| < \epsilon  $
\State $\quad \quad \mbox{Converged = TRUE}$
\State  $\quad \emph{else} \quad k = k+1$
\BState \emph{Output}:$\;  x$
\end{algorithmic}
\end{algorithm}

\begin{algorithm}[ht]
\caption{Cholesky Factor Estimation}\label{a:ecf}
\begin{algorithmic}[1]
\BState \emph{input}:
\State $\textit{$k_{max}, \mathbf{X}, \lambda, \gamma, \epsilon$}$
\State $L^{(0)} \gets \textit{Initial Cholesky factor} $
\State $ \textit{For}\; $i = 1,2,\dots, p$  $
\BState $\quad \beta^{i} = \argmin_{\beta_i} Q_{RRCF,i}(\beta^i)\;\;\textit{via Algorithm~\ref{a:cca}}$
\State $\textit{Construct $L \in \mathcal{L}_p$ by setting its non-zero values as $\beta^i$}$
\BState \emph{Output}:$\; \textit{Lower diagonal matrix $L$}$
\end{algorithmic}
\end{algorithm}


\subsubsection{Convergence of the Cyclic Coordintewise Algorithm}
As discussed, the score function $Q_{RRCF}(L)$ is non-convex with respect to $L$, and the convergence of iterates in Algorithm~\ref{a:ecf} can be guaranteed only to a local minimum. Next lemma shows that for the fixed permutation matrix $P$, the objective function $Q_{RRCF}(L)$ is lower bounded, a local minimum lies in the space of lower triangular matrices $\mathcal{L}_p$ with positive diagonal entries, and for certain values of $\gamma$, the generic function $h(\cdot)$ is strictly convex.

 \begin{lemma} \label{l:convanal}
 \begin{itemize}
\item[a.] If $A_{ii} > 0$, for $1 \leq i \leq p$
\[h_{k,A,\lambda,\gamma}(x) \geq 2x_k - 2.\]
\item[b.] For $\gamma > \max\{1/2A_{ii},1\}$, $h_{k,A,\lambda,\gamma}(x)$ is a strictly convex function of $x_i$ for $1 \leq i \leq k - 1$.
\item[c.] For every $n$ and $p$
\[
\begin{aligned}
\inf_{L \in \mathcal{L}_p}Q_{RRCF}(L) &= \sum_{i =1}^p \inf_{\beta^i} Q_{RRCF,i}(\beta^i)\\
&\geq -2p > - \infty
\end{aligned}
\]
and any local minimum of $Q_{RRCF}$ over the open set $\mathcal{L}_p$ lies in $\mathcal{L}_p$.
\end{itemize}
\end{lemma}
From this lemma, we can establish the convergence of the cyclic coordintewise algorithm.
\begin{theorem}\label{t:conv}
Under assumptions of Lemma~\ref{l:convanal}, Algorithm~\ref{a:ecf} converges to a local minimum of $Q_{RRCF}(L)$.
\end{theorem}

\section{Simulation and Data Analysis}
In this section, we study the empirical performance of our estimator on simulated and macro-economic datasets. The simulation results indicate that for a fixed p, if the number of edges in the DAG increases, i.e., the DAG is denser, the performance of RRCF tends to improve. The macro-economic data analysis provides the application of RRCF to solve the price puzzle \cite{Sims1992}, a well known problem in economics.

\subsection{Simulation Study}

We compare the performance of our algorithm with the three recent BN learning algorithms: \textbf{ARCS}:\cite{ye2019}, \textbf{CCDr}:\cite{aragam2015}, and
\textbf{NOTEARS} \cite{zheng2018}.
The performance is measured, both in terms of the structure learning, and how well the weighted adjacency matrix $\hat B$ estimates $B$.

The weighted adjacency matrix $B$ is constructed following \cite[Section 4.1]{kalisch2008} framework. We adapt parameterization $L = \Omega^{-1/2}(I - B)$ to generate data, where $\Omega = I_p$. The dimension of the data varies $p \in \{100, 200\}$ and the expected sparsity levels are  $s \in \{p, 2p\}$. The latter corresponds to the expected number of edges in the DAG. In all simulations, the sample size is $n = 150$ and each sample follows $p$-dimensional normal distribution $N(0, (L^tL)^{-1})$. 
 Each of the simulation settings $(p,s)$ is repeated over 20 datasets. The tuning parameters $\eta, \mu, \lambda$ for the RRCF algorithm are selected using the extended BIC criterion \cite{foygel2010} over the specified grid (See Supplementary for details). We note that, for the case $p \approx n$, one computational disadvantage of RRCF, compare to CCDr or NOTEARS, is a need to tune three tuning parameters which can be computationally costly for high dimensional datasets.

\subsubsection{Structure Learning and Estimation Accuracy}

We compare the four algorithms using the following four metrics: True Positive Rate (TPR), False Positive Rate (FPR), Structural Hamming Distance (SHD), and scaled Frobenius norm, which estimates how far the weighted adjacency matrix $\hat B$ is from $B$; i.e.,  $\frac{1}{p}\|\hat B - B\|_F$.

\begin{table}[ht!]
\centering
\caption{Average of three metrics over 20 replication for four $(p,m)$ settings. For TPR, a larger value indicates better performance; for FPR and FRB. NORM, a smaller value indicates better performance.}
\label{t:res1}
\setlength\extrarowheight{-3pt}
\begin{tabular}{c|c|c|c|c}
\hline
\rule{0pt}{12pt}
$(p,s)$ &Method & TPR & FPR & FRB. NORM\\
\hline
\rule{0pt}{12pt}
\multirow{4}{*}{ (100,100)}&ARCS & 0.601 & 0.001 & 7.052\\
&CCDr & \textbf{0.621} & \textbf{0.001} & 9.930\\
&RRCF & 0.603 & 0.001 & \textbf{6.868}\\
&NOTEARS & 0.612 & 0.001 & 9.930\\
\hline
\rule{0pt}{12pt}
\multirow{4}{*}{(100,200)}&ARCS & 0.637 & \textbf{0.003} & \textbf{10.569}\\
&CCDr & 0.636 & 0.005 & 14.305\\
&RRCF & \textbf{0.649} & 0.009 & 10.599\\
&NOTEARS & 0.645 & 0.004 & 11.349\\
\hline
\rule{0pt}{12pt}
\multirow{4}{*}{(200,200)}
&ARCS & 0.611 & \textbf{0.001} & 12.883\\
&CCDr & 0.651 & 0.007 & 19.845\\
&RRCF & 0.623 & 0.003 & 12.509\\
&NOTEARS & \textbf{0.657} & \textbf{0.001} & \textbf{12.233}\\
\hline
\rule{0pt}{12pt}
\multirow{4}{*}{(200,400)}
&ARCS & 0.635 & 0.001 & 13.043\\
&CCDr & 0.658 & \textbf{0.001} & 17.679\\
&RRCF & 0.643 & 0.001 & \textbf{12.752}\\
&NOTEARS & \textbf{0.655} & 0.002 & 14.617\\
\hline
\end{tabular}%
\end{table}

\begin{figure}[ht!]
\centering
\includegraphics[width= 6.5cm, height = 5.8cm]{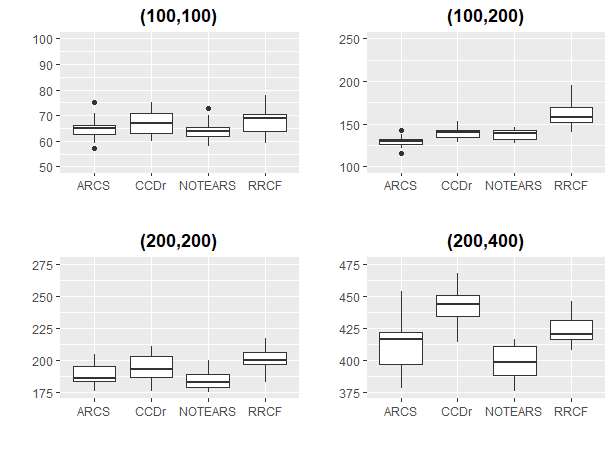}
\caption{Structural Hamming Distance boxplot for four $(p,s)$ settings. A smaller SHD value indicates better performance.}
\label{fig:shd}
\end{figure}

Table~\ref{t:res1} and Figure~\ref{fig:shd} report the simulation results. The best average score for each metric and  $(p,s)$ setting is highlighted in bold. Results suggest that RRCF performance improves when $s$ is higher for fixed $p$. In particular, for the $(100,100)$ case, CCDr provides the best results for the TPR and FPR metrics, followed by NOTEARS and RRCF. The situation changes for the $(100, 200)$ case, where RRCF provides the best TPR average score, and ARCS provides the best FPR average score. NOTEARS perform the best when the dimension increases from the 100 to 200. RRCF provides the best scaled Frobenius norm result for $(100,100)$ and $(200, 200)$ settings.

From Figure~\ref{fig:shd}, overall, the performance of RRCF is compatible with the considered algorithms.

\subsection{Macro-Economic Application}

We illustrate the application of the RRCF algorithm to the macro-economic dataset. In particular, we utilize our methodology to estimate the contemporaneous causal influences in the structural vector autoregression (SVAR) model. Then utilize impulse response functions to analyze the dynamics in SVAR models \cite[Section 2.3.2]{Lutkepohl2007} and discover the future effects of a shock on variables.

For a $K \times 1$ vector $Y_t = (y_{1t}, \dots, y_{Kt})^t$ the SVAR with $p_0$ lags is defind as 
\begin{equation}
Y_t = B_0 Y_t  +\sum_{i=1}^{p_0}B_iY_{t-i} + \varepsilon_t,
\end{equation} 
where $E(\varepsilon_t \varepsilon^t_t) = \Omega$ is a diagonal matrix and $B_i$'s  are $K \times K$ matrices. Notice that the relations among the contemporaneous components of $Y_t$ are embedded in the matrix $B_0$ and such causal structure can be represented by a DAG, whose vertices are the elements of the vector $Y_t$. That is there is a directed edge pointing from $y_{it}$ to $y_{jt}$ if and only if $(B_0)_{ij} \neq 0$. The  knowledge of the ordering among contemporaneous error terms is used for the estimation of the impulse response functions (for details, see \cite[Section 2]{hyvarinen2010}).  \cite{bessler1998, Demiralp2003} exploit PC algorithm for Gaussian data and \cite{hyvarinen2010, moneta2013, dallakyan2020} propose methods for non-Gaussian data to learn the contemporaneous ordering. \cite{chu2008, malinsky2018} propose methods for settings with unmeasured confounding. 
Algorithm~1 in \cite{dallakyan2020} summarizes steps on the use of DAGs for the SVAR estimation. We iterate it in Algorithm~\ref{alg:dagsvar} by incorporating the RRCF step in line 7 to recover the ordering of error terms.

\begin{algorithm}[!ht]
\caption{SVAR procedure with RRCF} \label{alg:dagsvar}
\begin{algorithmic}[1]
\Procedure{}{}
\BState \emph{input}:
\State $\textit{$y_1,\dots,y_t$} \gets \text{$K$ dimensional }\textit{stationary series}$
\BState \emph{top}:
\State   \textbf{Estimate} \text{the VAR model} \textit{$y_t = A_1y_{t-1}+ \dots + A_p y_{t-p} + u_t$, 
\State   \textbf{Estimate}  \textit{the residuals $\hat u_t = y_t -  \hat A_1y_{t-1}- \dots - \hat A_p y_{t-p}$}
\State   \textbf{Perform} \textit{ RRCF algorithm on residuals to recover ordering among residuals} $\hat u_{1t}, \dots, \hat u_{Kt}$}.
\BState \emph{Output}:
\State $B_0$
\EndProcedure
\end{algorithmic}
\end{algorithm}

\begin{figure}[ht!]
\centering
\begin{subfigure}{.25\textwidth}
  \centering
  \includegraphics[width=.8\linewidth, height = 3cm]{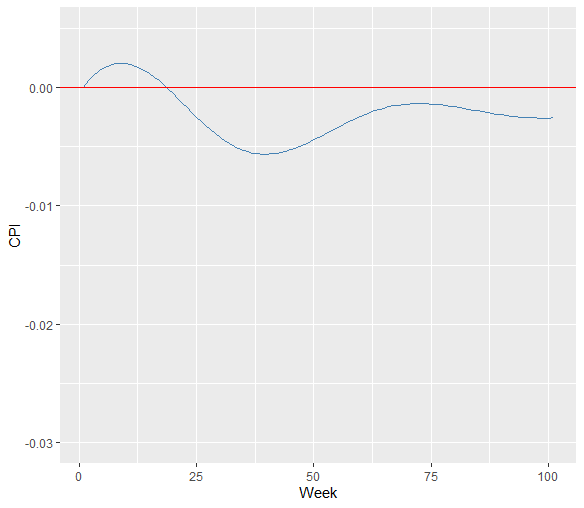}
  \caption{BA}
  \label{fig:FFRirfPC}
\end{subfigure}%
\begin{subfigure}{.25\textwidth}
  \centering
  \includegraphics[width=.8\linewidth, height = 3cm]{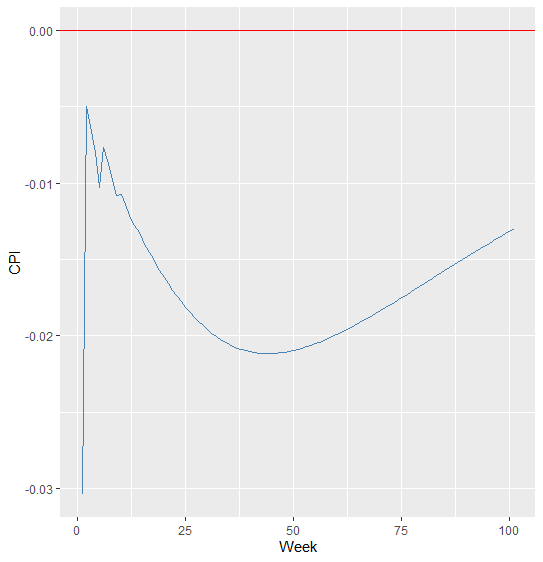}
  \caption{RRCF}
  \label{fig:FFRirf}
\end{subfigure}
\caption{The response of the Consumer Price Index to the Federal Fund Rate shock.}
\label{fig:svarirf}
\end{figure}

We use RRCF incorporated Algorithm~\ref{alg:dagsvar} to solve the price puzzle. The price puzzle in a structural autoregression (SVAR) system is known as an inability to explain the positive relationship between an innovation(shock) in the federal funds rate (FFR) and inflation \cite{Bernanke1992,Sims1992,Balke1994,Hoover2014}. It is a puzzle since an increase in the federal funds is expected to be followed by a decrease in the price level rather than an increase (See Figure~\ref{fig:FFRirfPC}).

\cite{dallakyan2020} showed that utilization of the recent DAG techniques to recover the ordering of error terms in VAR mitigates the price puzzle problem. Here, we show that using the sparse VAR approach and RRCF algorithm to recover the DAG structure of error terms leads to the complete disappearance of the price puzzle  (See Figure~\ref{fig:FFRirf}).

To analyze the price puzzle, we use a relatively rich dataset from \cite{Hoover2014}. Data consist of 12 monthly series for the United States that run from 1959:02 to 2007:06. Data sources and details are provided in \cite{Hoover2014}.
In the dataset, monetary policy is represented both by the Federal funds rate (FFR) and two reserve components: (the logarithms of) borrowed reserves (BORRES) and nonborrowed reserves (NBORRES). Financial markets are represented by two monetary aggregates (the logarithms of) M1 and (the non-M1 components of) M, as well as by three interest rates: the own-rate of interest on M2 (M2OWN), the 3-month Treasury bill rate (R3M), and the 10-year Treasury bond rate (R10Y ). Prices are represented by (the logarithms of) the consumer price index (CPI) and an index of sensitive commodity prices (COMPRICE). Finally, the real economy is represented by the (logarithm of) industrial production (INDPRO) and the output gap (GAP). Our sample period runs from January 1990 until 2009. The sample period is chosen such that to avoid a policy break \cite{Hoover2014}.


To introduce sparsity in the VAR estimation, we impose a lasso penalty on the VAR coefficient matrix \cite{Song2011,Nicholson2016}. For the sparse VAR estimation, we use the \texttt{BigVAR} package in \texttt{R} \cite{bigvar}, with the number of lags equal to 4. Then, using contemporaneous time restrictions obtained from Algorithm~\ref{alg:dagsvar}, estimate impulse response functions. For comparison, we include the impulse response function obtained from the procedure proposed in \cite{bessler1998}(BA for short).

Figure~\ref{fig:svarirf} plots the responses of Consumer Price Index (CPI) to Federal Fund Rate (FFR) obtained from the BA and RRCF algorithms, respectively. From Figure~\ref{fig:FFRirfPC}, the prize puzzle is apparent when the response is estimated using the BA's procedure. However, it disappears  when the response is estimated using the RRCF algorithm (see Figure~\ref{fig:FFRirf}). The latter result is consistent with the macro-economic literature.

\section{Statistical Properties} \label{s:statprop}
In this section, we study the consistency of the RRCF estimator, assuming that the true permutation matrix $P$ is known; i.e., data have known order. Under this assumption, the dependence of $Q_{RRCF}$ on $P$ is omitted and focus is only on the consistency of a Cholesky factor estimator in (\ref{eq:frrcf}).

\cite{khare2016, yu2017} provide consistency of the sparse Cholesky factor estimator for the convex objective function. However, our objective function is \textit{non-convex} and it may possess multiple local optima that are not global. Therefore, the standard statistical techniques are not applicable for establishing consistency.

We establish upper bounds on the Frobenius norm between \textbf{any local optimum} of the empirical estimator and the unique minimizer of the population. Even though the non-convex function may possess multiple local optima, our theoretical results guarantee that, from a statistical perspective, all local optima are  fundamentally as good as a global optimum. The theoretical analysis relies on the following assumptions: 
\begin{itemize}
\item{A1} \textit{Marginal sub-Gaussian assumption:} The sample matrix $X \in \mathcal{R}^{n \times p}$ has $n$ independent rows with each row drawn from the distribution of a zero-mean random vector $X = (X_1, \dots, X_p)^t$ with covariance $\Sigma$ and sub-Gaussian marginals; i.e.,
\[E[\mbox{exp}(tX_j/ \sqrt{\Sigma_{jj}})] \leq \mbox{exp}(Ct^2) \]
for all $j=1,\dots,p, \, t \leq 0$ and for some constant $C > 0$.
\item{A2} \textit{Sparsity Assumption:} The true Cholesky factor $L \in \mathcal{R}^{p \times p}$ is the lower triangular matrix with positive diagonal elements and support $\mathcal{S}(L) =\{(i,j), i \neq j| L_{ij} \neq 0\} $. We denote by $s = |S|$ cardinality of the set $S$.
\item{A3} \textit{Bounded eigenvalues:} There exist a constant $\kappa$ such that
\[0 < \kappa^{-1} \leq \lambda_{min}(L) \leq \lambda_{max}(L) \leq \kappa \]
\end{itemize}
Before providing our main result, we recall that a matrix $\hat L \in \mathcal{L}_p$ is a stationary point for $Q_{RRCF}$ if it satisfies \cite{bertsekas2015}
\begin{equation} \label{eq:stat}
\langle\nabla \mathcal{L}_n(\hat L) + \nabla \rho(\hat L, \lambda), L - \hat L \rangle \geq 0,\; \mbox{for}\, L \in \mathcal{L}_p,
\end{equation}
where $\mathcal{L}_n(L) = \mbox{tr}(SLL^t) - 2\log|L|$ and $\nabla \rho(\cdot,\cdot)$ is the subgradient.
\begin{theorem} \label{t:stprop}
Under Assumptions A1-A3, with tuning parameter $\lambda$ of scale $\sqrt{\frac{\log p}{n}}$, and $\frac{3}{4 \gamma} < (\kappa + 1)^{-2}$, the scaling $(s+p)\log p = o(n)$ is sufficient for any stationary point $\hat L$ of the non-convex program $Q_{RRCF}$ to satisfy the following estimation bounds:
\[
\begin{aligned}
\|\hat L - L\|_F &= \mathcal{O}_p \Big (\sqrt{\frac{(s + p)\log p}{n}} \Big ) \\
\|\hat \Sigma^{-1} - \Sigma^{-1}\|_F &= \mathcal{O}_p \Big (\sqrt{\frac{(s + p) \log p}{n}} \Big ) \\
\end{aligned}
\]
\end{theorem}
The proof is provided in Supplementary.

\section{Conclusion}
The present paper proposes two-step algorithm to learn a DAG from the regularized score function when data are generated from a Gaussian, linear SEM. The first step introduces a permutation matrix as a new parameter to represent  variable ordering.
For its estimation, we utilize a relaxation technique in which we relax the non-convex space of permutation matrices by the convex space of doubly stochastic matrices. Then choose the ``closest'' permutation matrix to the optimal doubly stochastic matrix.
 In the second step, given the variable ordering, the algorithm estimates a Cholesky factor, which entails the DAG structure. For each step, we provide necessary conditions that guarantee convergence of the proposed algorithm. The numerical results study the advantage and potential applications of the algorithm in recovering DAG structure.

As future work, we left the proof of the statistical consistency of the RRCF estimator when the permutation matrix $P$ is unknown, selection of the penalty parameter $\mu$ when $n <<p$, and the possible extension of the proposed method to learn DAGs from the non-linear SEMs.

\ifCLASSOPTIONcaptionsoff
  \newpage
\fi



\bibliographystyle{IEEEtran}
\bibliography{chol_bib}
\end{document}


%
\title{Supplementary for Learning Bayesian Networks through Birkhoff Polytope: A Relaxation Method}
%
%
%

\author{Aramayis~Dallakyan,
        Mohsen~Pourahmadi
\thanks{Department of Statistics, Texas A\&M University, College Station, TX, 77843}
}

\maketitle


%
\IEEEpeerreviewmaketitle

\section{Proofs} \label{a:proof}

\subsection*{Proof of Proposition~\ref{p:prop1}}
Before delving into details, we define two important concepts that will be used in the proof: majorization and Schur-convexity. For details, see \cite{marshall11}.

\begin{definition}
	For $x,y \in \mathcal{R}^p$, $x$ is said to be majorized by $y$
	
	\[ x \prec y \; \mbox{if}\, \begin{cases} \sum_{i =1}^k x_{[i]} \leq \sum_{i=1}^{k} y_{[i]}, & k=1,\dots p-1 \\
									\sum_{i=1}^p x_{[i]} = \sum_{i = 1}^p y_{[i]},
	\end{cases}\]
	where $x_{[1]} \geq \dots \geq x_{[n]}$ denotes the components of $x$ in decreasing order.
\end{definition}

\begin{definition}
A real-valued function $f$ defined on a set $\mathcal{A} \subset \mathcal{R}^p$ is said to be Schur-convex on $\mathcal{A}$ is
\[x \prec y \; \mbox{on} \; \mathcal{A} \Rightarrow f(x) \leq  f(y) \]
\end{definition}

Now, since from \cite[page 20]{marshall11} a Schur-convex function is permutation invarient, we only need to show that $\ell(L,I| \mathbf{X})$ is Schur-convex. That is we show if $\mathbf{x} = \mbox{vec}(\mbox{X}) \prec \mathbf{y} = \mbox{vec}(\mbox{Y})$ then $\ell(L,I| \mathbf{X}) \leq \ell(L,I| \mathbf{Y})$, where for the $n \times p$ matrix $X$,  the $np \times 1$ vector $ \mbox{vec}(X)$ is its standard vectorization formed by stacking up its column vectors.

From (\ref{eq:parloglik}),
\[\ell(L,I| \mathbf{X}) = \mathbf{x}^t (L^tL \otimes I) \mathbf{x} + C,\]
where $C$ contains terms constant with respect to $X$. Thus, to proof the Schur-convexity,  we need to show that $\mathbf{x}^t (L^tL \otimes I) \mathbf{x}  \leq \mathbf{y}^t (L^tL \otimes I) \mathbf{y}$, which easily follows from the \cite[Chapter 4, Proposition B.9]{marshall11}.
\subsection*{Proof of Lemma~\ref{l:dsprop}}

For the proof of only if side; i.e., $P$ is a permutation matrix or $P = J/p$, the equality holds trivially from the definition of the Frobenius norm \cite[Chapter 5.6]{Horn2012}.

For the if part, it is known that the maximum spectral radius and the spectral norm $\|P\|_2$ of the doubly stochastic matrix $P \in \mathcal{D}_p$ are equal 1 \cite[Chapter 8.7]{Horn2012}. From the matrix norm inequality \cite[Corollary 5.4.5]{Horn2012}

\begin{equation} \label{eq:frbds}
1 = \|P\|^2_2 \leq \|P\|^2_F =\sum_{j = 1}^{p}\sigma^2_j(P)  \leq p \|P\|^2_2 \leq p,
\end{equation}
where $\sigma_j(\cdot)$ is the $j$th singular value. Thus, $\|P\|_F = \|P\|_2 = 1$ if and only if it is a matrix of rank one; i.e., $P = J/p$ from the \cite[Theorem 4]{marcus1962}.

On the other hand, it is easy to see that $\|P\|^2_F = p$ equality holds if and only if $\sigma_j(p) = 1,\, j= 1, \dots, p$, that is $P \in \mathcal{P}_p$, and any doubly stochastic matrix  with  Frobenius norm $\sqrt{p}$  is a permutation matrix.
%


\subsection*{Proof of Corollary~\ref{c:optsol}}
We write for $P \in \mathcal{D}_p$
\[ 
\begin{aligned}
    \mbox{tr}(LPSP^tL^T) & = \mbox{vec}(P)^t(L^tL \otimes S) \mbox{vec}(P)\\
    &\geq \lambda_1(L^tL \otimes S)\|P\|^2_F,
\end{aligned}
\]
where $\lambda_1(A)$ is the smallest eigenvalue of the symmetric matrix $A$  , $\mbox{vec}(\cdot)$ is the usual matrix vectorization operator and the minimum value achieved when $P = J/p$ from the Lemma~\ref{l:dsprop}.

\subsection*{Proof of Lemma~\ref{l:l1}}
\paragraph*{(a):} The Hessian of the objective function can be found by noting that
\[ 
\begin{aligned}
    \frac{1}{2} \mbox{tr}(LPSP^tL^t)  - \frac{1}{2}\mu\|P\|^2_F = &\frac{1}{2}\mbox{vec}(P)^t (L^tL \otimes S) \mbox{vec}(P)\\
    &- \frac{1}{2} \mu \mbox{vec}(P)^t \mbox{vec}(P).
\end{aligned}
\]
Thus the Hessian is
\[L^tL \otimes S - \mu \mathbf{I},\]
and the result easily follows from the definition of convexity \cite{Boyd2004}.
\paragraph*{(b):} We first show that the transformation (\ref{p:trstoc}) is equivalent to (\ref{p:stocort}). Following \cite{fogel2013}, we write
\[\|TP\|^2_F = \mbox{tr}(P^tT^tTP) = \mbox{tr}(P^tP - \mathbf{1}\mathbf{1}^t/p) = \|P\|^2_F - 1, \]
where we use idempotent property of the projection matrix $T$. Thus,  (\ref{p:trstoc}) has the same objective function as (\ref{p:stocort}) up to a constant.
To show convexity we look on the Hessian of the objective function. Note that
\[
\begin{aligned}
    \frac{1}{2}\mbox{tr}(LPSP^tL^t)  - \frac{1}{2}\mu\|TP\|^2_F &= \frac{1}{2}\mbox{vec}(P)^t (L^tL \otimes S) \mbox{vec}(P) \\
    &- \frac{1}{2} \mu \mbox{vec}(P)^t(\mathbf{I} \otimes T) \mbox{vec}(P), 
\end{aligned}
\]
where $vec(\cdot)$ is the usual matrix vectorization operator. Thus the Hessian is
\begin{equation} \label{e:hess}
L^tL \otimes S - \mu \mathbf{I} \otimes T
\end{equation}
and under $\mu \leq \lambda_2(S)\lambda_1(L^tL)$ convexity holds.

\paragraph*{(b):}
Similarly, from (\ref{e:hess}) if $\mu > \lambda_m(S)\lambda_m(L^tL)$ then the objective function of (\ref{p:trstoc}) is concave. Thus, from the  \cite[Corrollary 8.7.4]{Horn2012} the minimum of concave function over the set of doubly stochastic matrices is attained at a permutation matrix and the proof follows.

\subsection*{Proof of Lemma~\ref{l:lbound}}
We start by assuming contradiction. The proof exploits strategy introduced in \cite{rothman2008}.  Let for $\tilde P \in \mathcal{D}_p$ belonging to a Birkhoff polytope
\begin{equation} \label{e:lbound1}
\begin{aligned}
Q(\tilde P) &= \mbox{tr}(L \tilde P S \tilde P^t L^t) - \frac{1}{2} \mu \|\tilde P\|^2_F \\
               & - \frac{1}{2} tr(LPSP^tL^t) - \frac{1}{2} \mu \|P\|^2_F\\
                &=\mbox{tr}(L(\tilde P - P)S(\tilde P - P)^t L^t)\\
                &- \frac{1}{2}\mu(\|\tilde P\|^2_F - \|P\|^2_F),
\end{aligned}
\end{equation}
where $P \in \mathcal{P}_p$ is a true permutation matrix, $\|P\|^2_F = p$ from Lemma~\ref{l:dsprop}.
Our estimate $\hat P$ minimizes $Q(\tilde P)$, or equivalently $\hat \Delta = \hat P - P$ minimizes $G(\Delta) = Q(P + \Delta)$, where
 $\Delta = \tilde P - P$. Under convexity condition in Lemma~\ref{l:l1}(a), $G(\Delta)$ is a convex function and $G(\hat \Delta) \leq G(0) = 0$.

Next we introduce the set
\[\Theta_n = \{\Delta: \|\Delta\|_F = r_n\},\]
where $r_n \rightarrow 0$.
Now if we show that $\inf \{G(\Delta): \Delta \in \Theta_n\} >0$  then $\hat \Delta \in \Theta_n$ and $\|\hat \Delta\|_F \leq r_n$.

We write
\begin{equation} \label{e:lbound2}
G(\Delta) = I + II,
\end{equation}
where $I = \mbox{tr}(L \Delta S \Delta^t L^t)$ and $II = \frac{\mu}{2} (\|P\|^2_F - \|P + \Delta\|^2_F)$.

For the part $I$,
 \begin{equation} \label{e:lbound5}
 \begin{aligned}
       I = \mbox{tr}(L^tL \Delta S \Delta^t) &= \mbox{vec}(\Delta)^t (L^tL \otimes S) \mbox{vec}(\Delta)\\
       &\geq \lambda_1(L^tL \otimes S) \| \Delta \|^2_F
 \end{aligned}
 \end{equation}



 To find lower bound for $II$, we denote by $C = \{(i,j):P_{ij} = 1\}$ non-zero entries of the permutation matrix $P$. We also note that the cardinality $|C| = p$. Thus,

 \[ 
 \begin{aligned}
     \|P + \Delta\|^2_F &= \sum_{(i,j) \in C}|1 + \Delta_{ij}|^2 + \sum_{(m,n) \not \in C} |\Delta_{mn}|^2\\ 
     &= p + 2 \sum_{(i,j) \in C}|\Delta_{ij}| + \|\Delta\|^2_F
 \end{aligned}
  \]
 and
 \begin{equation} \label{e:lbound7}
 \begin{aligned}
      \|P + \Delta\|^2_F - \|P\|^2_F &=  2 \sum_{(i,j) \in C}|\Delta_{ij}| + \|\Delta\|^2_F \\
      & \leq 2p \max_{i,j}|\Delta_{ij}| +  \|\Delta\|^2_F\\
      &\leq 2p +  \|\Delta\|^2_F ,
 \end{aligned}
  \end{equation}
 where we used the norm inequality $\|A\|_{max} \leq \|A\|_F$ and $\max_{i,j}|\Delta_{ij}| \leq 1$.
 It follows from (\ref{e:lbound5}) and \ref{e:lbound7}  the lower bound for (\ref{e:lbound2}) is

  \begin{equation} \label{e:lbound8}
 \begin{aligned}
 	G(\Delta)  & \geq  \lambda_1(L^tL \otimes S) \|\Delta\|^2_F - 2p \mu - \mu \|\Delta\|^2_F \\
 		       & = \|\Delta\|^2_F  ( \lambda_1(L^tL \otimes S)  - \mu - \frac{2p}{r^2_n}\mu)
\end{aligned}
 \end{equation}

 Thus, $G(\Delta) > 0$ condition holds if and only if
  \[ \lambda_1(L^tL \otimes S)   - \mu - \frac{2p}{r^2_n}\mu > 0, \]
 from which follows that
 \[\mu < \frac{ \lambda_1(L^tL \otimes S) }{1 + \frac{2p}{r^2_n}}\]
and from $r_n \rightarrow 0$, it follows $\mu \rightarrow 0$, which contradicts the initial statement.



\subsection*{Proof of Lemma~\ref{l:xstar}}
We start by writing for $1 \leq j \leq  k -1$
\begin{equation} \label{eq:objfncj}
h_{k,A,\lambda,\gamma} = x_j^2A_{jj} + 2x_j (\sum_{l \neq j} A_{lj}x_l) + \rho(|x_j|, \lambda,\gamma) + C_j,
\end{equation}
where $C_j$ includes terms independent of $x_j$. Taking derivative with respect to $x_j$ and noting that the subdifferential
 \begin{equation}\label{eq:mcp}
  \rho^{'}(|x_j|, \lambda, \gamma) = \begin{cases} \lambda s - \frac{x_j}{\gamma} & |x_j| < \gamma \lambda \\
  0 & |x_j| \geq \gamma \lambda
  \end{cases}
  \end{equation}
Here, the subgradient $s = \mbox{sign}(x_j)$ if $x_j \neq 0$ and take values in $[-1,1]$ otherwise. Thus it follows
\[x^*_j = \frac{S_\lambda(-2\sum_{l \neq j} A_{lj}x_l)}{2A_{jj} - 1/\gamma}\]

Similarly,
\[h_{k,A,\lambda,\gamma} = x_k^2A_{kk} + 2x_k (\sum_{l \neq k} A_{lj}x_l) -2\log x_k+ C_k,\]
where $C_k$ includes terms independent of $x_k$ and after taking derivatives with respect to $x_k$
\[
\begin{aligned}
    &\frac{-2}{x_k} + 2x_kA_{kk} + 2 \sum_{l \neq k}A_{lk}x_l = 0 \iff \\
   & x^2_k A_{kk} + \sum_{l \neq k}A_{lk}x_l x_k -1 =0,
\end{aligned}
  \]
and (\ref{eq:xk}) follows after retaining the positive root of the above quadratic equation.

\subsection*{Proof of Lemma~\ref{l:convanal}}

\paragraph*{(a):}
From (\ref{e:gen})
\[h_{k,A,\lambda,\gamma}(x) \geq 2x_k - 2,\] 
where we used that $A$ is positive semidefinite and $x^tAx \geq 0,\; \rho(|x_i|, \lambda, \gamma)  \geq 0$ and for $y > 0$, $\log y \leq 1 - y$.

\paragraph*{(b):}
We denote by $D_{u}h$ and $D^2_{u}h$ the derivative and the second derivative of $h$ in the direction of $u$. Thus, the proof follows from (\ref{eq:objfncj}) and (\ref{eq:mcp}) by writing

\[\min \{D^2_{\beta^{-}_{i}}h_{k,A,\lambda,\gamma}(\beta),D^2_{\beta^{+}_{i}}h_{k,A,\lambda,\gamma}(\beta)\} \geq 2A_{ii} - \frac{1}{\gamma} \]

\paragraph*{(c):}
Note that
\[\inf_{L \in \mathcal{L}_p}Q_{RRCF}(L)  \geq -2p > - \infty\]
directly follows from the part (b) of the proof. Moreover, from (\ref{eq:rrcf1}) if $|\beta^i_j| \rightarrow \infty$ or $\beta^i_i = 0$, then $Q_{RRCF} \rightarrow \infty$. Therefore, any local minimum of $Q_{RRCF}$ over the open set $\mathcal{L}_p$ lies in $\mathcal{L}_p$.

\subsection*{Proof of Theorem~\ref{t:conv}}
For the proof of Theorem~\ref{t:conv}, we exploit \cite[Theorem 5.1]{tseng2001}, where the author established sufficient conditions for the convergence of cyclic coordinate descent algorithms to coordinate-wise minima. The strict convexity of (\ref{eq:rrcfi}) with respect to each coordinate direction and lower boundedness established in Lemma~\ref{l:convanal} satisfy the required conditions in Theorem 5.1. Thus, convergence to a coordinate-wise minimum point is guaranteed. Moreover, since all directional derivatives exist, every coordinate-wise minimum is also a local minimum.

\subsection*{Conditions for the penalty function} \label{pr:condF}
 The penalty function $\rho(\cdot, \lambda)$ satisfies the following conditions:
\begin{itemize}
\item The function $\rho(\cdot, \lambda)$ satisfies $\rho(0,\lambda) = 0$ and is symmetric around zero.
\item On the non negative real line, the function  $\rho(\cdot, \lambda)$  is nondecreasing.
\item For $t > 0$, the function $t \rightarrow \rho(\cdot, \lambda) / t$ is nonincreasing in $t$.
\item The function $\rho(\cdot, \lambda)$ is differentiable for all $t \neq 0$ and subdifferentiable at $t = 0$, with $\mbox{lim}_{t \rightarrow 0^+} \rho'(t, \lambda) = \lambda C$.
\item There exists $\mu > 0$ such that $\rho_{\mu}(t, \lambda) = \rho(t, \lambda) + \frac{\mu}{2}t^2$ is convex.
\end{itemize}

\subsection*{Proof of Theorem~\ref{t:stprop}}
We start by showing that $\mathcal{L}_n$ satisfies RSC conditions. Recall that the differentiable function $\mathcal{L}_n: \mathcal{R}^{p \times p} \rightarrow \mathcal{R}$ satisfies RSC condition if:
\[
\langle \nabla \mathcal{L}_n(L + \Delta) - \nabla \mathcal{L}_n(L), \Delta \rangle \geq
\]
\begin{numcases}{\geq}
 	\alpha_1 \| \Delta \|^2_F - \tau_1 \frac{\log p}{n} \| \Delta \|^2_1, & $\forall \| \Delta \|_F \leq 1$  \label{eq:rsc1}\\
        \alpha_2 \| \Delta \|_F - \tau_2 \sqrt{\frac{\log p}{n}} \| \Delta \|_2, & $\forall \| \Delta \|_F \geq 1$  \label{eq:rsc2}
\end{numcases}
 where the $\alpha_j$'s are strictly positive constants and the $\tau_j$'s are nonnegative constants. From \cite[Lemma 4]{loh2015}, under conditions of Theorem~\ref{t:stprop}, if (\ref{eq:rsc1}) holds then (\ref{eq:rsc2}) holds. Thus, we concentrate only on showing that  (\ref{eq:rsc1}) holds for $\| \Delta \|_F \leq 1$.
 Recall that
 \begin{equation} \label{e:ln}
 \mathcal{L}_n(L) = \mbox{tr}(SL^tL) - 2\log|L|
\end{equation}

\begin{lemma} \label{l:rsc}
 The cost function (\ref{e:ln}) satisfies RSC condition (\ref{t:stprop}) with $\alpha_1 = (\kappa + 1)^{-2}$ and $\tau_1 = 0$; i.e.,
 \begin{equation} \label{eq:rscp}
  \langle \nabla \mathcal{L}_n(L + \Delta) - \nabla \mathcal{L}_n(L), \Delta \rangle \geq  (\kappa + 1)^{-2} \| \Delta \|^2_F, \: \forall \| \Delta \|_F \leq 1
 \end{equation}
\end{lemma}
The proof is provided in Appendix~\ref{a:rsc}.

From the penalty conditions listed above, $\rho_{\mu}(L,\lambda) = \rho(L, \lambda) + \frac{\mu}{2} \|L\|^2_F$ is convex, where in case of MCP $\mu = 1/ \gamma$. Thus,
\[
\begin{aligned}
\rho_{\mu}(L,\lambda) - \rho_{\mu}(\hat L,\lambda) &\geq \langle \nabla \rho_{\mu}(\hat L,\lambda), L - \hat L \rangle \\
&=\langle \nabla \rho(\hat L,\lambda) + \mu \hat L, L - \hat L \rangle, 
\end{aligned}
\]
which implies that
\begin{equation} \label{eq:rho}
 \langle \nabla \rho( \hat L,\lambda), L - \hat L \rangle \leq \rho( L,\lambda) - \rho( \hat L,\lambda) + \frac{\mu}{2} \|\hat L - L\|^2_F
\end{equation}

From stationarity condition (\ref{eq:stat})
\[\langle \nabla \mathcal{L}_n (\hat L), L - \hat L \rangle \geq - \langle \nabla \rho(\hat L, \lambda), L - \hat L \rangle\]
and combining above result with (\ref{eq:rscp})
\[ \begin{aligned}
(1 + \kappa)^{-2}\|\Delta\|^2_F &\leq \langle \mathcal{L}_n(\hat L), \Delta \rangle - \langle \nabla \mathcal{L}_n(L), \Delta \rangle \\
& \leq \langle \nabla \rho(\hat L, \lambda), L - \hat L \rangle - \langle \nabla \mathcal{L}_n(L), \Delta \rangle  \\
& \leq \rho(L, \lambda) - \rho(\hat L, \lambda) + \frac{\mu}{2} \|\hat L - L\|^2_F  \\
&- \langle \nabla \mathcal{L}_n(L), \Delta \rangle
\end{aligned}
\]
After rearrangement and H{\''o}lder inequality
\[
\begin{aligned}
\Big((1 + \kappa)^{-2}) - \frac{\mu}{2} \Big) \|\Delta\|^2_F & \leq \rho(L, \lambda) - \rho(\hat L, \lambda)\\ & + \|\nabla \mathcal{L}_n(L)\|_{\infty} \|\Delta\|_1
\end{aligned}
\]
From \cite[Lemma 4]{loh2015}
\[\lambda \|\Delta\|_1 \leq \rho(\Delta,\lambda) + \frac{\mu}{2}\|\Delta\|^2_F\]
and from \cite[Lemma 15]{yu2017} under the assumed scaling of $\lambda$
\[\|\nabla \mathcal{L}_n(L)\|_{\infty} \leq \frac{\lambda}{2}\]
with probability going to 1. Combining above two results and using subadditive property; i.e., $\rho(\Delta,\lambda) \leq \rho(L, \lambda) + \rho(\hat L, \lambda)$:
\[
\begin{aligned}
\Big( (1 + \kappa)^{-2} - \frac{\mu}{2} \Big ) \|\Delta\|^2_F &\leq \rho(L, \lambda) - \rho (\hat L, \lambda) + \frac{\lambda}{2} \|\hat \Delta\|_1\\
& \leq \rho(L, \lambda) - \rho (\hat L, \lambda) \\
&+ \frac{\rho (\Delta, \lambda)}{2} + \frac{\mu}{4} \|\Delta\|^2_F \\
& \leq \rho(L, \lambda) - \rho(\hat L, \lambda)\\
&+ \frac{\rho (L, \lambda) + \rho(\hat L, \lambda)}{2} + \frac{\mu}{4} \|\Delta\|^2_F
\end{aligned}
 \]
After rearranging and using $3/4\mu \leq (1+ \kappa)^{-2} $
\begin{equation} \label{eq:cnd1}
0 \leq \Big( (1 + \kappa)^{-2} - \frac{3}{4}\mu \Big) \|\Delta\|^2_F \leq 3 \rho(L, \lambda) - \rho(\hat L, \lambda)
\end{equation}
From (\ref{eq:cnd1}) and \cite[Lemma 5]{loh2015} follows
\[ \rho(L, \lambda) - \rho(L, \lambda) \leq 2\lambda \|\Delta_S\| - \lambda \|\Delta_{S^c}\| \Rightarrow \|\Delta_{S^c}\|_1 \leq 3 \|\Delta_S\|_1\]
Thus,
\[
\begin{aligned}
\Big(2 (1 + \kappa)^{-2}  - \frac{3}{2}\mu \Big) \|\Delta\|^2_F  &  \leq \lambda \|\Delta_S\|_2 - \lambda \|\Delta_{S^c}\|_1 \\
& \leq \lambda \|\Delta_S\|_1 \leq \lambda \sqrt{p + s} \|\Delta\|_F,
\end{aligned}
\]
from which we conclude that
\begin{equation} \label{eq:res1}
\|\Delta\|_F \leq \frac{6\lambda \sqrt{p + s}}{4(1 + \kappa)^{-2} - 3 \mu},
\end{equation}
and the result follows from the chosen scaling of $\lambda$.

For the precision matrix bound, from page 45 of \cite{yu2017} we note that
\[  \hat L^t \hat L - L^tL = (\hat L - L)^t (\hat L - L) + (\hat L - L)^TL + L^t (\hat L - L)\]
and
\[\|L^t (\hat L - L)\|_F \leq |||L|||_2 \|\hat L - L\|_F\]
From submultiplicativity property of matrix norm
\[\|(\hat L - L)^t (\hat L - L)\|_F \leq \|(\hat L - L)\|^2_F \]
Therefore,
\begin{equation} \label{eq:res2}
 \|\hat L^t \hat L - L^tL\|_F \leq (\|\hat L - L\|_F + 2 |||L|||_2)\|\hat L - L\|_F
 \end{equation}

\subsection*{Proof of Lemma~\ref{l:rsc}} \label{a:rsc}
The following facts will be useful in the proof.
\paragraph*{Fact 1}
\begin{enumerate}
\item $(K_{pp})^{-1} = K_{pp}$
\item $\lambda_{max}(K_{pp}) = 1$
\item $tr(ABCD) = vec(D^t)(C^t \otimes A)vec(B)$
\item $\lambda_{max}(A \otimes B) = \lambda_{max}(A)\lambda_{max}(B)$
\end{enumerate}
where $K_{pp}$ is the commutation matrix  such that $vec(L) = K_{pp} vec(L^t)$. The proof of the facts can be found in \cite[Section 4]{magnus1986}.

To show the RSC condition, we rely on the directional derivatives (for example see \cite[Section 6.3]{tao2016}). In particular, if we denote by $D_{\Delta}\mathcal{L}_n(L)$ the directional derivative with respect to the direction $\Delta$, then from \cite[Lemma 6.3.5]{tao2016} :
\begin{equation} \label{e:dder1}
\langle \nabla \mathcal{L}_n(L), \Delta \rangle = D_{\Delta}\mathcal{L}_n(L) = 2\mbox{tr}[(SL^t - L^{-1})\Delta]
\end{equation}
Similarly
\begin{equation} \label{e:dder2}
\begin{aligned}
\langle \nabla \mathcal{L}_n(L + \Delta), \Delta \rangle &= D_{\Delta}\mathcal{L}_n(L + \Delta) \\
&= 2\mbox{tr}[(S(L+\Delta)^t - (L+ \Delta)^{-1})\Delta]
\end{aligned}
\end{equation}
From Woodbury identity \cite{Horn2012}
\[(L + \Delta)^{-1} = L^{-1} - L^{-1}\Delta(L + \Delta)^{-1}\]
Plugging back into (\ref{e:dder2}) and after some algebra
\begin{equation} \label{e:dder2.1}
\begin{aligned}
\langle \nabla \mathcal{L}_n(L + \Delta), \Delta \rangle &= 2\mbox{tr}[(S(L+\Delta)^t - (L+ \Delta)^{-1})\Delta] \\
&+ 2\mbox{tr}[S \Delta^t\Delta + L^{-1}\Delta(L + \Delta)^{-1}\Delta]
\end{aligned}
\end{equation}
Thus, from (\ref{e:dder1}) and (\ref{e:dder2.1})

\begin{equation} \label{eq:dder3}
\begin{aligned}
\langle \nabla \mathcal{L}_n(L + \Delta) &- \nabla \mathcal{L}_n(L), \Delta \rangle =\\ &= 2\mbox{tr}[\Delta^tS\Delta + L^{-1}\Delta(L+\Delta)^{-1}\Delta] \\
& \geq \mbox{vec}(\Delta)^t K_{pp}((L + \Delta)^{-t}\otimes L^{-1}) \mbox{vec}(\Delta)\\
& = \mbox{vec}(\Delta)^t [((L + \Delta)^{t}\otimes L)K^{-1}_{pp}]^{-1} \mbox{vec}(\Delta)\\
& \geq \lambda_{min}([((L + \Delta)^{t}\otimes L)K^{-1}_{pp}]^{-1})\|\Delta\|^2_F,
\end{aligned}
\end{equation}
where for the first inequality we used the fact that $S$ is positive semi-definite and the second equality follows from the Fact 1.
Now, since
\begin{equation} \label{eq:preres}
\begin{aligned}
\lambda_{min}([((L + \Delta)^{t}&\otimes L)K^{-1}_{pp}]^{-1}) =\\
&= \lambda^{-1}_{max}[((L + \Delta)^{t}\otimes L)K^{-1}_{pp}] \\
&\geq \lambda^{-1}_{max}(K^{-1}_{pp}) \lambda^{-1}_{max}(L) \lambda^{-1}_{max}(L + \Delta)\\
&\geq (\kappa + 1)^{-2} ,
\end{aligned}
\end{equation}
where the first inequality follows from the submultiplicativity property of the norm and lower-triangularity of the $L$ and $\Delta$. The second inequality follows from the triangular property, the fact that $\|\Delta\|_2 \leq \|\Delta\|_F \leq 1$ and, properties of the $K_{pp}$ stated in Fact 1. After plugging (\ref{eq:preres}) into (\ref{eq:dder3}), the result follows.

 \section{Algorithms and related derivations} \label{a:alg}

\subsection*{Algorithm and derivation of the closed form solution for itarates in  (\ref{p:dual})} \label{a:dual}
For each iteration $k$, we use the following notation $ f(P^{(k)}) = \frac{1}{2} \|P^{(k)} - P_0\|$, and  $f_*(u^{(k)},v^{(k)},U^{(k)}) =  -\frac{1}{2}\|u^{(k)}\mathbf{1}^t + \mathbf{1}(v^{(k)})^t - U^{(k)}\|^2_F - tr((U^{(k)})^tP_0) $.
\begin{algorithm}[H]
\caption{Projection on doubly stochastic matrices}\label{a:dsm}
\begin{algorithmic}[1]
\BState \emph{input}:
\State $\textit{$k_{max}, \epsilon$} \gets \textit{max. number of iteration and stopping criteria}$
\State $\textit{$U^{(0)} \in R^{p \times p}_{+}, u^{(0)},v^{(0)} \in R^{p}$} \gets \textit{Initial dual variables}$
\State $\textit{converged = FALSE}$
\BState \emph{while converged == False $\;\textit{and}\; k < k_{max}$}: 
\State $\quad U^{(k)} \gets \max \{0, u^{(k-1)}\mathbf{1}^t + \mathbf{1}(v^{(k-1)})^t - P_0\}$
\State $\quad u^{(k)} \gets \frac{1}{p}(P_0 \mathbf{1} - ((v^{(k-1)})^t \mathbf{1}+ 1)\mathbf{1} + U^{(k)} \mathbf{1}) $
\State $\quad v^{(k)} \gets \frac{1}{p}(P^t_0 \mathbf{1} - ((u^{(k)})^t\mathbf{1} + 1)\mathbf{1} + (U^{(k)})^t \mathbf{1})$
\State   $\quad  \textit{Update primal variable:\;} P^{(k)} = P_0 - u^{(k)}\mathbf{1}^t - \mathbf{1}(v^{(k)})^t + U^{(k)}$
\State $\quad \textit{If}\;  |f(P^{(k)}) - f_*(u^{(k)},v^{(k)},U^{(k)})| < \epsilon$
\State $\quad \quad \textit{converged = TRUE}$
\State $\quad \textit{else}$
\State  $ \quad \quad k = k+1$
\BState \emph{Output}:$\; \textit{Doubly Stochastic Matrix}\; P $
\end{algorithmic}
\end{algorithm}

The \textbf{convergence of Algorithm~\ref{a:dsm} is guaranteed} since the objective function is differentiable and strictly concave in each block component when all other block components are held fixed \cite[Proposition 6.5.2]{bertsekas2015}.

\subsection*{Dual function derivation}
The Lagrangian of (\ref{p:proj}) is  \cite{bertsekas2015}
\[
\begin{split}
\mathcal{L}(P,u,v,U) &= \frac{1}{2}\|P - P_0\|^2_F + u^t (P \mathbf{1} - \mathbf{1} )\\
&+ v^t(P^t \mathbf{1} - \mathbf{1})  - tr(U^t P)
\end{split}\]
The dual objective function is defined as
\begin{equation} \label{e:dual}
\mathcal{L}_{*}(u,v,U) = \inf_{P}\mathcal{L}(P,u,v,U)
\end{equation}

Thus,
\[ P = P_0 - u \mathbf{1}^t - \mathbf{1}v^t + U \]

Plugging this back into (\ref{e:dual})
\[ \begin{aligned}
\mathcal{L}_{*}(u,v,U) &= \frac{1}{2}\|u\mathbf{1}^{t} + \mathbf{1}v^{t} - U\|^{2}_F + u^{t} (P_0 \mathbf{1} -\mathbf{1})\\
&+ v^t(P_0^t \mathbf{1} - \mathbf{1}) - tr(U^{t}P_0) \\
				  & -\mbox{tr}(u^{t}u\mathbf{1}^{t}\onevec  + u^{t}\onevec v^{t} \onevec + u^{t}U\onevec + v^{t}\onevec u^{t}\onevec \\
				  &+ v^{t}v\onevec^{t}\onevec + v^{t}u^{t}\onevec  \\
				  & - U^{t}u\onevec^t - U^{t}u\onevec^t - U^t \onevec v^t - U^{t}U)
\end{aligned}
\]
Then the (\ref{p:dual}) follows by noting that the expression in the trace function is equal to $\|u\mathbf{1}^t - \mathbf{1}v^t + U\|^2_F$. Now taking derivative with respect to $u,v, U$, the corresponding expressions for iterations in the Algorithm~\ref{a:dsm} follows.

\section{Tuning parameter selection} \label{a:ebic}

We use extended BIC (eBIC) criterion \cite{foygel2010} for the tuning parameters $\theta = (\lambda, \gamma)$ selection in Algorithm~\ref{a:RRCF}.   Ideally, we want to tune the parameters for each update of $\hat L^{(k)}$ in line 6, however, when the convergence takes more iteration this approach is computationally costly. In practice, the tuning parameters are selected before starting the iterations \cite{ye2019}, hence a particular scoring function is fixed throughout the algorithm.

The eBIC criterion takes the form

\[\mbox{BIC}_{\gamma_{BIC}}(\mathcal{S}(L)) = -2 \mathcal{L}_n(\hat L) + s\log n + 4 s\gamma_{bic} \log p,\]
where $\mathcal{S}(L)$ is the support of matrix $L$, $s = |\mathcal{S}(L)|$, $\mathcal{L}_n(\hat L)$ is the maximized log-likelihood function and $\gamma_{BIC} \in [0,1]$. A larger value of $\gamma_{BIC}$ results to a stronger penalization of $L$, and the case $\gamma_{BIC} = 0$ corresponds to the classical BIC.  In general, the value of $\gamma_{eBIC}$ is unknown, but relying on simulation results, authors suggest $\gamma_{eBIC} = 0.5$ as a candidate value.

%
%
%
%

\ifCLASSOPTIONcaptionsoff
  \newpage
\fi



\bibliographystyle{IEEEtran}
\bibliography{chol_bib}
%



%



